%% file: top.tex
\documentclass[10pt,twocolumn,letterpaper]{article}

\usepackage{cvpr}
\usepackage{times}
\usepackage{epsfig}
\usepackage{graphicx}
\usepackage{amsmath,amsfonts,amssymb,amsthm} % define this before the line numbering.
\usepackage[pagebackref=true,breaklinks=true,letterpaper=true,colorlinks,bookmarks=false]{hyperref}
\usepackage{color}
\usepackage{amsfonts}
\usepackage[subrefformat=parens,labelformat=parens]{subfig} 
\usepackage[linesnumbered,ruled,noend]{algorithm2e}
\usepackage{color}
\usepackage{xspace}
\usepackage{ifthen}
\usepackage{cite}

\definecolor{gray}{rgb}{0.5,0.5,0.5}

\SetCommentSty{mycommfont}

\input{defs}

\cvprfinalcopy 

\begin{document}
\title{FollowMe: Efficient Online Min-Cost Flow Tracking\\with Bounded Memory and Computation}

\author{Philip Lenz\\
Karlsruhe Institute of Technology\\
% Karlsruhe, Germany\\
{\tt\small lenz@kit.edu}
\and
Andreas Geiger\\
MPI T\"ubingen\\
% T\"ubingen, Germany\\
{\tt\small andreas.geiger@tue.mpg.de}
\and
Raquel Urtasun\\
University of Toronto\\
% Toronto, Canada\\
{\tt\small urtasun@cs.toronto.edu}
}

\maketitle

\input{abstract}

\input{intro}
\input{related}
\input{approach}
\input{dynamic}
\input{online}
\input{results}
\input{conc}

{\small
\bibliographystyle{ieee}
\bibliography{bibliography_long,bibliography}
}
\end{document}

%% file: defs.tex
\definecolor{forestgreen}{rgb}{0.0, 0.0, 0.0}

\newcommand{\shortgets}{\hspace*{-0.1em} \gets \hspace*{-0.1em}}

\DeclareMathOperator*{\argmin}{argmin}
\DeclareMathOperator*{\cost}{cost}

\newcommand{\comment}[1]{}

\newcommand{\tmax}{{t}} % number of frames
\newcommand{\bx}{\mathbf{x}}

\newcommand{\bs}{\mathbf{s}}

\newcommand{\bw}{\mathbf{w}}

\newcommand{\bff}{\mathbf{f}}

\newcommand{\bo}{\mathbf{o}}

\newcommand{\co}{{\cal O}}

\newcommand{\cc}{\mathcal{C}}

\newcommand{\cT}{{\cal T}}

\newcommand{\figref}[1]{Fig.~\ref{#1}}
\newcommand{\subfigref}[1]{Fig.~\subref*{#1}}
\newcommand{\secref}[1]{Section~\ref{#1}}
\newcommand{\algref}[1]{Algo.~\ref{#1}}
\renewcommand{\eqref}[1]{Eq.~\ref{#1}}

%% file: abstract.tex
\begin{abstract}
One of the most popular approaches to multi-target tracking is tracking-by-detection. Current  min-cost flow algorithms which solve the data association problem optimally have three main drawbacks: they are computationally expensive, they assume that the whole video is given as a batch, and they scale badly in memory and computation with the length of the video sequence. In this paper, we address each of these issues, resulting in a computationally and memory-bounded solution. First, we introduce a dynamic version of the successive shortest-path algorithm  which solves the data association problem optimally while reusing computation, resulting in significantly faster inference than standard solvers. Second, we address the optimal solution to the data association problem when dealing with an incoming stream of data (i.e., online setting). Finally, we present our main contribution which is an approximate online solution with bounded memory and computation which is capable of handling videos of arbitrarily length while performing tracking in real time. We demonstrate the effectiveness of our algorithms on the KITTI and PETS2009 benchmarks and show state-of-the-art performance, while being significantly faster than existing solvers. 
\end{abstract}

%% file: intro.tex
\section{Introduction}

In recent years the performance of object detectors has improved substantially, reaching levels which nowadays make them applicable to real-world problems. A notable example is face recognition where algorithms that employ deep learning features have attained human performance \cite{Taigman2014CVPR}.
Many aspects have been key for the success of object recognition.  The creation of benchmarks such as PASCAL \cite{Everingham2010IJCV} and ImageNet \cite{Russakovsky2014ARXIV} made experimentation more rigorous and stimulated the development of key algorithms such as the deformable part-based model (DPM) \cite{Felzenszwalb2010PAMI} and more recently deep learning approaches based on  convolutional nets such as \cite{Krizhevsky2012NIPS}  and R-CNN \cite{Girshick2014CVPR}. 

With the success of object detection came also the success of {\it tracking-by-detection} approaches \cite{Zhang2008CVPR,Pirsiavash2011CVPR,Andriyenko2012CVPR,Breitenstein2011PAMI,Leibe2007CVPR,Yang2012CVPR,Milan2013CVPR,Butt2013CVPR,Segal2013ICCV} which track an unknown number of objects over time. The idea is to first detect objects independently in each frame, followed by an association step linking individual detections to form trajectories. The association problem is  difficult due to the presence of occlusions, noisy detections as well as false negatives. 
The simplest approach is to independently establish associations between all pairs of consecutive frames, a problem which can be solved optimally in polynomial time using the Hungarian method. However, in challenging situations ambiguities are hard to resolve using local information alone and early errors cannot be corrected at later frames leading to globally inconsistent tracks. More recently, sophisticated discrete-continuous optimization schemes have been proposed \cite{Andriyenko2012CVPR}, which alternate between optimizing trajectories and performing data association over the whole sequence by assigning detections to the  set of trajectory hypotheses. Unfortunately, the resulting problems are highly non-convex, hard to optimize and provide no guarantees of optimality.

In their seminal work, Zhang \etal \cite{Zhang2008CVPR} showed how multi-frame data association can be cast as a network flow problem. 
The optimal solution can be found by a min-cost flow algorithm, solving simultaneously for the number of objects and their trajectories according to a cost-function which incorporates the likelihood of a detection as well as pairwise relationships between detections for all consecutive frames. Recent extensions \cite{Berclaz2011PAMI,Pirsiavash2011CVPR,Butt2013CVPR,Arora2013ICCV} demonstrate speedups by using more elaborate graph solvers, and show how higher-order motion models can be incorporated into the formulation at the cost of giving up global optimality.

In this paper we are interested in making min-cost flow solutions applicable to real-world scenarios such as autonomous driving. Towards this goal two main problems need to be addressed:  (i) current approaches assume a batch setting, where all detections must be available a priori, and  (ii) their memory and computation requirements grow unbounded with the size of the input video.
As a consequence existing min-cost flow techniques cannot be  employed in robotics applications where tracking algorithms are required to run non-stop in an online setting while competing with other processes for a limited computation and memory budget. 
In this paper we develop practical tracking by detection solutions that 
\begin{itemize}
\item perform {\it computations only when necessary},
\item handle an  {\it online} stream of data, and
\item are {\it  bounded in memory and computation}.
\end{itemize}
In particular,  we show that the first two properties can be achieved while maintaining  optimality, and an approximate solution is possible for the latter which still performs very well in practice. 
We demonstrate the effectiveness of the proposed solvers on the KITTI \cite{Geiger2012CVPR} and  PETS 2009 \cite{Ferryman2009PETS} tracking benchmarks. As evidenced by our experiments, our approach is significantly more efficient  than existing min-cost flow algorithms. Furthermore, a near-optimal solution is retrieved when bounding the memory and computation to as little as $20$kB and $10$ms. 
Our code will be made available upon publication.

%% file: related.tex
\section{Related Work}

Multi-target tracking approaches  can be divided into two main categories: filter\-ing-based approaches and batch methods. 
{\it Filtering-based methods} \cite{Mitzel2012ECCV,Breitenstein2011PAMI,Ess2009PAMI,Nam2014ECCV} are based on the Markov assumption, \ie, the current state depends only on the previous states. 
While they are fast and applicable in real-time applications, they typically suffer from their inability to recover from early errors.

Recent efforts focus primarily on {\it batch methods}, where object hypothesis are typically obtained using an object detector (tracking-by-detection) and  tracking is formulated as an optimization problem over the whole sequence. This mitigates the aforementioned problems and allows for the integration of higher-level cues such as social group behavior in the case of pedestrian tracking \cite{Choi2012ECCV,Kooij2012ECCV,Leal-Taixe2011ICCVWORK,LealTaixe2014CVPR,Chen2014CVPR,Shi2014CVPR}. In \cite{Andriyenko2011CVPR}, the task is formulated as a continuous energy minimization problem which allows incorporating richer motion models but is difficult to optimize. Discrete-continuous optimization techniques have been proposed \cite{Andriyenko2012CVPR,Milan2013CVPR}, where discrete (data association) and continuous (trajectory fitting) optimization problems are solved in an alternating way until converging to a hopefully better local minimum. 
In \cite{Benfold2011CVPR,Brau2013ICCV,Choi2013PAMI,Yang2014CVPR,Collins2014ECCV}, approximate Markov chain Monte Carlo techniques are employed for solving the data association problem. In order to increase the discriminative power of appearance and dynamical models, online learning approaches have been suggested \cite{Leal-Taixe2012CVPR,Li2009CVPRa,Yang2011CVPR,Song2010ECCV,Yang2012CVPR,Yang2012ECCV,Yu2013CVPR,Zhang2013CVPR}. 
In \cite{Berclaz2011PAMI,Shitrit2013PAMI}, the problem is cast as optimization on a grid using a linear program while assuming that the observing camera is static.
The problem of tracking through occlusions has been tackled in \cite{Idrees2012ECCV,Xiong2012ECCV,Xing2009CVPR,Liu2013CVPR,Dicle2013ICCV,Chen2014CVPR,Bae2014CVPR,Wen2014CVPR,Possegger2014CVPR} by using context from outside the object region or by building strong statistical motion models.

While most of the aforementioned formulations resort to approximate optimization without optimality guarantees,
Zhang \etal \cite{Zhang2008CVPR} showed how data association with pairwise energies can be formulated as a network flow problem such that standard graph solvers can be leveraged to retrieve the global optimum. Their formulation solves for the globally optimal trajectories including their number, and hence implicitly solves the model selection problem.
To reduce the computational complexity of min-cost flow algorithms,  \cite{Berclaz2011PAMI,Pirsiavash2011CVPR} proposed to use the successive shortest-path (SSP) algorithm as solver. Further speed-ups have been achieved in \cite{Pirsiavash2011CVPR} using a greedy dynamic programming (DP) approximation.
The min-cost flow idea has been extended in \cite{Butt2013CVPR,Arora2013ICCV} to include higher-order terms  at the price of loosing optimality. More recently, Wang \etal\, \cite{Wang2014ECCV} proposed to jointly model the appearance of interacting objects by integrating linear flow constraints into the framework.

While all of the existing min-cost flow formulations assume a batch setting, in this paper we propose a computationally and memory-bounded version of this algorithm, which is able to process video sequences frame-by-frame while reusing computation via efficient caching strategies. Furthermore, our scheduling strategy performs computations only when necessary which also speeds up traditional batch solvers.

%% file: approach.tex
\section{Review on Optimal Tracking-by-Detection} \label{sec:review}

One of the most popular approaches to multi-target tracking is tracking-by-detection, where a set of detections are computed for each frame and trajectories are formed by linking these detections. 
In this section, we briefly review the necessary preliminaries to our contributions in \secref{sec:dynamicOnlineBounded}. In particular, we review how to formulate multi-target tracking as a min-cost flow problem and how to solve it using the successive shortest path algorithm.

\subsection{Tracking as Min-Cost Flow}

Following the tracking-by-detection paradigm, we assume that a set of detections ${\cal X} = \{\bx_i\}$ is available as input. 
Let $\bx_i = (x_i,t_i,s_i,a_i,d_i)$ denote a detection, with $x_i$ the position of the bounding box, $t_i$ the time step (frame index), $s_i$ the size of the bounding box, $a_i$ the appearance, and $d_i$ the detector score. 
We define a trajectory as a sequence of observations $T_k = (o_1, o_2, \ldots,  o_{l_k})$ with $o\in\{1,\ldots,|\cal X|\}$ the detection index of  temporally adjacent detections $t_{o_{i+1}} = t_{o_i}+1$.  

The full association hypothesis is then given by a set of trajectories ${\cal T} = \{T_k\}$, and the 
 data association problem can be formulated as a Markov random field (MRF). More specifically, we aim at maximizing the posterior probability of trajectories:
\begin{equation}
p({\cal T}|{\cal X})  = p(\cT) \, \prod_i p(\bx_i|{\cal T}) 
\label{eq:prob}
\end{equation}
The observation model is given by
\begin{equation}
p(\bx_i|{\cal T})  = \left\{
\begin{array}{ll}
P_i &\mbox{ if $\exists \,T_k \in {\cal T} \land i \in T_k$} \\
1-P_i &\mbox{ otherwise}
\end{array} \right.
\end{equation}
where $P_i$ denotes the probability of $\bx_i$ being a true detection. % and $1-\beta_i$ is the probability of a false alarm.
The prior over trajectories decomposes into a product of unary and pairwise factors
\begin{equation}
p({\cal T}) \propto \prod_{T \in {\cal T}} \Psi(T) \, \prod_{T,T' \in {\cal T}} [T \cap T' = \emptyset]
\end{equation}
where the pairwise term ensures that trajectories are disjoint. The unary factors are given by
\begin{equation}
\Psi(T) = \Psi_{en}(\bx_{o_1}) \Psi_{ex}(\bx_{o_l}) \prod_{i=1}^{l-1} \Psi_{li}(\bx_{o_i},\bx_{o_{i+1}})
\end{equation}
where $\Psi_{en}(\bx_{o_1})$, $\Psi_{ex}(\bx_{o_l})$ and $\Psi_{li}(\bx_{o_i},\bx_{o_{i+1}})$  model the likelihood of entering a trajectory, exiting a trajectory and linking  temporally adjacent  detections within a trajectory.

Taking the negative logarithm of (\eqref{eq:prob}), the maximization can be transformed into an equivalent minimization problem over flow variables \cite{Zhang2008CVPR} as follows 
 \begin{align}
{\cal \bff}^* = \argmin_{ \bff}\,
   &\sum_i C^{en}_{i} f^{en}_i + \sum_i C^{ex}_i f^{ex}_i \nonumber \\ + &  \sum_{i,j} C^{li}_{i,j} f^{li}_{i,j} + \sum_i C^{det}_{i} f^{det}_{i} \nonumber \\
  s.t. \;\, f^{en}_i + &\sum_j f^{li}_{j,i} = f^{det}_i = f^{ex}_i + \sum_j f^{li}_{i,j} \quad \forall i
  \label{eq:flow}
 \end{align}
where $C^{en}_i = -\log\Psi_{en}(\bx_i)$ is the cost of creating a new trajectory at $\bx_i$ and  $C^{ex}_i=-\log \Psi_{ex}(\bx_i)$ is the cost of exiting a trajectory at $\bx_i$. The cost of linking two consecutive  detections $\bx_i$ and $\bx_j$ is denoted $C^{li}_{i,j} = -\log \Psi_{li}(\bx_i,\bx_j)$ and $C^{det}_i$ encodes the cost of $\bx_i$  being a true detection or a false positive (data term). 
Furthermore, the binary flow variables $f^{en}_i$ or $f^{ex}_i$ take value 1 if the solution contains a trajectory such that $\bx_i$ is the first frame or last frame, respectively. $f^{det}_i=1$ encodes the fact that $\bx_i$ is part of a trajectory and $f^{li}_{i,j}=1$ if a tracklet exists which contains both detections $\bx_i$ and $\bx_j$ in two consecutive frames. 

\begin{figure}[t!]
  \centering
  \begin{minipage}[b]{0.9\linewidth}
    \includegraphics[width=\linewidth]{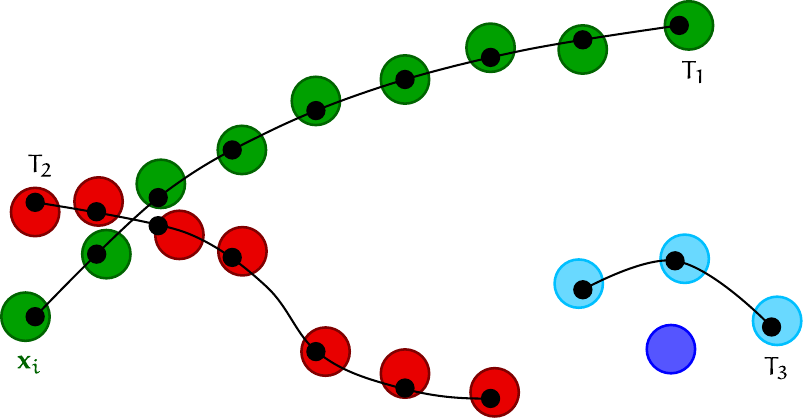}
  \end{minipage}\\
\vspace{0.2cm}
  \begin{minipage}[b]{0.9\linewidth}
    \includegraphics[width=\linewidth]{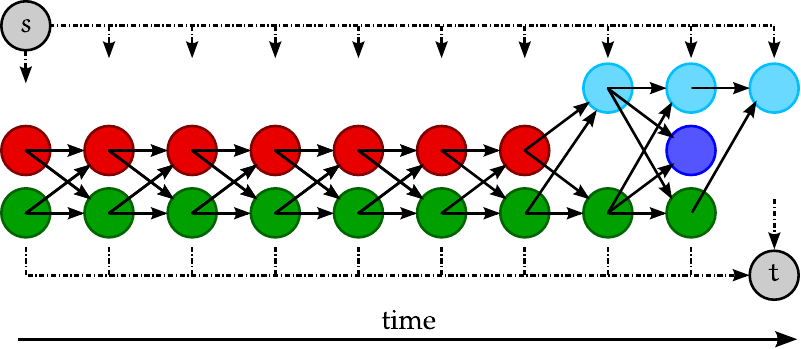}
  \end{minipage}
  \caption{The original problem (top) is mapped into a min-cost flow network (bottom). Ground truth trajectories are shown in black. Colored nodes encode detections and correspond to two nodes in the min-cost flow network $G$. For clarity of illustration, edges from the source and to the sink have been ommitted.}
  \label{fig:problem_mapping}
\vspace{0.3cm}
\end{figure}

In their seminal work, Zhang and Nevatia \cite{Zhang2008CVPR}  showed how to map the problem in \eqref{eq:flow} into a min-cost flow network problem. Fig. \ref{fig:problem_mapping} illustrates one such network graph, where for each observation two nodes $u_i$ and $v_i$ are created (summarized as one node for clarity of illustration) with an edge between them with cost $c(u_i,v_i)=C^{det}_i$ and flow $f(u_i,v_i) = f^{det}_i$.
For entering, edges are introduced between the source $s$ and the first node of each detection with cost $c(s,u_i)=C^{en}_i$ and flow $f^{en}_i$.
For exiting, edges are introduced  between the last node of each detection and the sink with cost $c(v_i,t)=C^{ex}_i$ and flow $f^{ex}_i$.
Finally, edges between consecutive detections $(v_i, u_j)$ encode pairwise association scores with cost $c(v_i,u_j) = C^{li}_{i,j}$ and flow $f^{li}_{i,j}$. 
While we assume that only detections in consecutive frames can be linked, this can be easily generalized by allowing transitions between detections in non-consecutive frames resulting in additional edges.
 
To find the optimal solution, Zhang \etal \cite{Zhang2008CVPR} start with flow zero, and augment the flow one unit at a time, applying the push relabeling algorithm \cite{Goldberg1997JA} to retrieve the shortest path at each iteration. The algorithm terminates if the cost of the currently retrieved shortest path is greater or equal to zero.
For efficiency, the bisection method can be applied on the number of flow units, reducing time complexity from linear to logarithmic with respect to the number of trajectories. The total complexity is then  ${\cal O}((mn^2\log^2(n)))$, with $n$ the number of nodes and $m$ the number of edges. 

\subsection{Successive Shortest-Path (SSP)} % for Min-Cost Flow
\label{subsec:SSP}

Recently, \cite{Berclaz2011PAMI,Pirsiavash2011CVPR}  proposed to replace the costly push relabeling algorithm by the successive shortest-path (SSP) algorithm \cite[p.~104]{Ahuja1993}. This reduces the computational complexity to ${\cal O}(K(n\log(n)+m))$ where $n$ is the number of nodes, $m$ is the number of edges and $K$ denotes the number of trajectories which is upper bounded by the number of nodes $n$. This section gives a brief introduction to the SSP algorithm as it forms the foundation for our contributions.  We refer the reader to the supplementary material for technical details in greater depth and an illustrative example.

The SSP algorithm  works as follows: it first computes the shortest path between source and target (\ie, path with  the lowest negative cost). 
It then iterates between reversing the edges of the previously found shortest path to form the residual graph $G_r$,  and computing the shortest path in this new residual graph.   This process is iterated until no trajectory with negative cost can be found.  
 Finally, trajectories are extracted by backtracking connected, inverted edges starting at the target node. 

In the first iteration the shortest path from the source to the target is efficiently retrieved using dynamic programming \cite[p.~655]{Cormen2001} as the input graph is directed and acyclic. In the following iterations, Dijkstra's algorithm can be leveraged after converting the graph such that all costs are positive.
This conversion is achieved by simply replacing the current cost $C_{u,v}$ between nodes $(u,v)$ linked by a directed edge, with $C'_{u,v} = C_{u,v} + d(u) -d(v)$, where $d(u)$ and $d(v)$ are the distance on the shortest path from the source to nodes $u$ and $v$ (\cf supplementary material). Importantly, for $u$ and $v$ on the shortest path we have $C'_{u,v} = 0$ after conversion.
Note that for graphs with positive costs and unit flow capacity, the SSP algorithm is similar to the k-shortest paths (``KSP'') algorithm \cite{Suurballe1974Networks,Suurballe1984Networks}. 
Furthermore, one could use other algorithms such as  Bellman-Ford to replace Dijkstra and handle  negative costs directly. However,  this  would result in worst complexity, \ie, $\co(K(n^2))$ . 

%% file: dynamic.tex
\section{Dynamic Online and Bounded Tracking}
\label{sec:dynamicOnlineBounded}

In this section we present new algorithms that have the necessary properties  for tracking-by-detection to be applicable to  real-world scenarios: dynamic computation, handling online data, and bounded memory and computation. We start by proposing a distributed optimal SSP algorithm which leverages a dynamic priority cue, saving computation when compared to the original SSP algorithm. 
We then extend this algorithm to handle an online data stream. Finally, we proposed our main contribution, a memory and computation bounded online SSP algorithm. 

\subsection{Optimal Dynamic Min-Cost Flow (dSSP)}
\label{subsec:dDijkstra}

\begin{figure*}[ht]
\subfloat[Update Nodes\label{fig:dDijkstra:a}]{\includegraphics[width=0.2\textwidth]{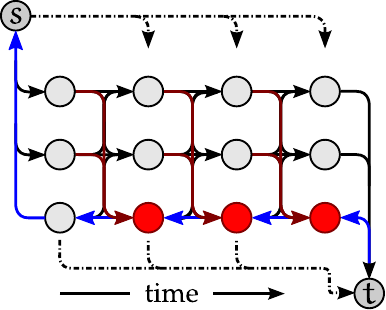}}
\subfloat[Backward Iteration\label{fig:dDijkstra:b}]{\includegraphics[width=0.2\textwidth]{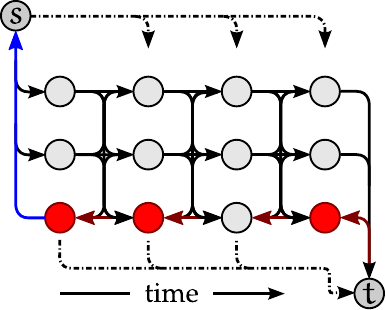}}
\subfloat[Forward Iteration\label{fig:dDijkstra:c}]{\includegraphics[width=0.2\textwidth]{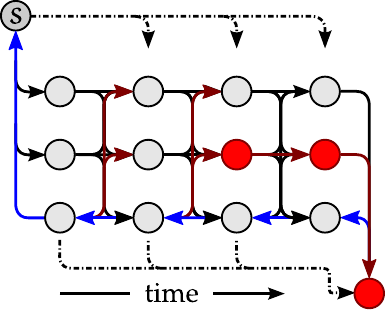}}
\subfloat[Shortest Path\label{fig:dDijkstra:d}]{\includegraphics[width=0.2\textwidth]{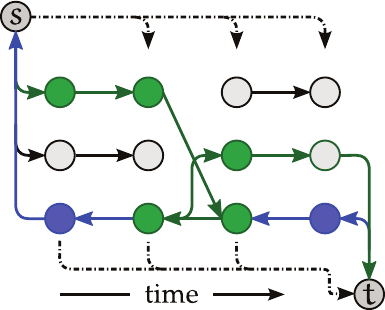}}
\subfloat[Final Trajectories\label{fig:dDijkstra:e}]{\includegraphics[width=0.2\textwidth]{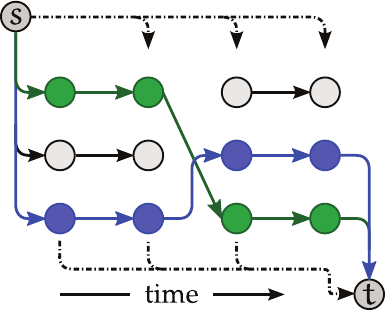}}
\caption{{\bf Dynamic Message Broadcasting (dSSP):}
For a given residual graph (shortest path in blue), nodes with invalid predecessors which require an update (red) are detected and queued \protect\subref{fig:dDijkstra:a}.
The queue is processed \protect\subref{fig:dDijkstra:b}, successively taking nodes from the queue and relaxing their outgoing edges. Successors with updated predecessors (red) are added to the queue \protect\subref{fig:dDijkstra:c}. Exploiting Dijkstra's algorithm, the priority queue is processed until no nodes are left. The algorithm terminates with the solution for the current residual graph \protect\subref{fig:dDijkstra:d}+\protect\subref{fig:dDijkstra:e}. Trajectories are encoded by backward pointing edges (green and blue).
}
\label{fig:dDijkstra}
\vspace{-0.3cm}
\end{figure*}

In this section we propose  a novel {\it dynamic} algorithm which performs computations only when necessary. 
Dijkstra's algorithm is based on the principle of relaxation, in which an upper bound of the correct distance is gradually replaced by tighter bounds (by computing the predecessor and its distance) until the optimal solution is reached. Nodes are held in a priority queue, guaranteeing that the most promising node is relaxed in each iteration \footnote{This is guaranteed as all costs are positive. 
}. This priority queue is initialized to hold only the source node, and upon convergence all nodes have been visited.

The  key intuition behind our dynamic computation is that calling Dijkstra at each iteration of the SSP algorithm is suboptimal in terms of runtime as from one iteration to the next only a small part of the graph has changed (\ie, only the  reversed edges are different). 
Inspired by the distributed Bellman-Ford algorithm 
\cite{Bertsekas1992,Walden2003}, we proposed to reuse computation and  only update the  predecessors when needed. This can be easily implemented within Dijkstra by using a {\it dynamic priority queue}, which contains only nodes that have changed.  
We refer the reader to \algref{algo:ssp_offline_dDijkstra} for a summary of our dynamic algorithm. 

\begin{figure}[t!]
\centering
\subfloat[Dijkstra (SSP)\label{fig:dDijkstra:h}]{\includegraphics[width=0.2\textwidth]{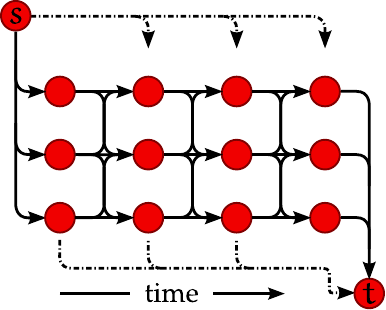}}
\qquad
\subfloat[Dynamic Dijkstra (dSSP)\label{fig:dDijkstra:f}]{\includegraphics[width=0.2\textwidth]{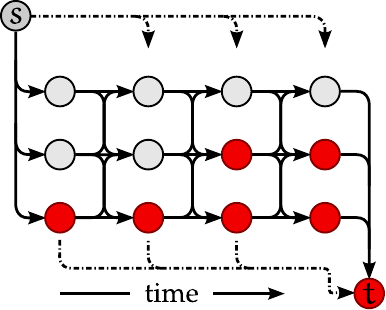}}
\caption{{\bf Message Broadcasting Strategies.} For Dijkstra's original algorithm the full predecessor map is computed by broadcasting messages to all nodes \protect\subref{fig:dDijkstra:h}. In contrast, our dynamic message broadcasting scheme exploits the intrinsic properties of the tracking problem and updates only deprecated predecessor maps using a priority cue \protect\subref{fig:dDijkstra:f}.}
\label{fig:dDijkstraComparison}
\end{figure}

We illustrate our dynamic algorithm  within the computation of a single iteration of SSP in \figref{fig:dDijkstra} for an example containing four frames with three detections each. 
Given the shortest path found in the previous iteration, we revert its edges to form the residual graph in \subfigref{fig:dDijkstra:a}. Note that the corresponding predecessor maps have to be updated as the direction of these edges has changed. 
We start by updating all predecessors for nodes belonging to the most recent trajectory (blue path in \subfigref{fig:dDijkstra:a}) in a forward sweep (relaxed edges are marked in red). Next, all nodes with a new predecessor propagate their cost along their respective shortest path to their successors when taken from the priority queue (\figref{fig:dDijkstra}~\subref{fig:dDijkstra:b},\subref{fig:dDijkstra:c}). 
In turn these successors are pushed into the priority queue if they received an update (\ie, the predecessor changed). 
The algorithm (one iteration of SSP) terminates when the queue is empty as all shortest paths have been updated.
Note that this dynamic computation is employed at each iteration of SSP, and  
SSP  terminates when no new shortest path with negative cost can be found (\subfigref{fig:dDijkstra:d}). The final trajectories are  extracted by collecting all backward pointing edges of $G_r$, starting at the target node (\subfigref{fig:dDijkstra:e}).
In contrast to Dijkstra's original algorithm, our dynamic broadcasting scheme relaxes only parts of the graph for typical tracking networks as illustrated in \figref{fig:dDijkstraComparison} and demonstrated later in our experimental evaluation.

\begin{algorithm}[t]
  \DontPrintSemicolon
  \KwIn{Set of Detections $\mathcal{X} = \{\mathbf{x_i}\} 
  $ }
  \KwOut{Set of trajectory hypotheses $\mathcal{T} = \{T_k\}\;$ 
  }
  $G(V,E,C,f) \gets$ ConstructGraph($\mathcal{X},s,t$)\;
  $f(G) \gets 0$ \tcp*{initialize flow to 0}
  $\gamma_0, \pi_0 \gets$ DAG-SP($G(f)$) \tcp*{shortest path in DAG}
  $G_r^{(0)}(f) \gets$ ConvertEdgeCosts($G(f), \gamma_0, \pi_0$) 
  $G^{(1)}_r(f) \gets$ ComputeResidualGraph($G^{(0)}_r(f),\pi^{(0)}$)\;
  $q \gets \emptyset$ \tcp*{$q$ is maintained for every iteration $k$}
  \While(\tcp*[f]{find shortest paths for $k\geq1$}){$1$} {
    $k \gets k+1$\;
%    \tcp{initial predecessor map $\pi_k$ (invalid)}
    $\pi^{(k)} \gets \pi^{(k-1)}$\;
%    \tcp{predecessors on $\gamma^{(k-1)}$ are invalid}
    $q \gets \gamma^{(k-1)}$\;
          \tcp{process queue in time direction}
    \ForEach{node $u \in q$}{
      \tcp{check predecessor from past}
      $\pi^{(k)} \gets$ Update($\pi^{(k)}$, $u$)\; 
      \If{updated}{
        $q \gets$ AddSuccessors($q$, $u$, $G^{(k)}_r(f)$)\;
      }
    }
    \tcp{process queue}
    \While{$\lnot$ empty($q$)}{
      $u \gets$ NodeWithMinDistance($q$)\tcp*{pop node}
      $q \gets q \setminus u$\; 
      %$S \gets S \cup \{u\}$\;
      \ForEach{node $v \in $ Successors($G^{(k)}_r(f)$,$u$)}{
        \tcp{Check predecessor of $v$}
        $\pi^{(k)}(v) \gets$ Relax($u$,$v$,$c$)\;
        \If{$d(v)>d(u)+c(u,v)$}{
          $d(v) \gets d(u)+c(u,v)$\;
          $q \gets$ AddNode($q$, $v$)\;
        }
      }
    }
    $G^{(k)}_r(f) \gets$ ConvertEdgeCost($G^{(k)}_r(f), \pi^{(k)}$)\;
    $G^{(k+1)}_r(f) \gets$ ResidualGraph($G^{(k)}_r(f), \gamma^{(k)}$)\;
    \tcp{evaluate converted costs}
    \If{$\sum_{i=1}^k \cost(\gamma^{(i)}) > |\cost(\gamma^{(0)})|$}{
      break\;
    }
  }
  \Return{$\mathcal{T}$}\;
  \caption{dSSP}
  \label{algo:ssp_offline_dDijkstra}
\end{algorithm}

%% file: online.tex
\subsection{Optimal Online Solution (odSSP)}\label{sec:online_optimal}

Next, we extend the dynamic min-cost flow formulation introduced in the previous section to the online setting.
Our intuition is that every time a new observation arrives (\ie, set of detections at time $t$), we would like to reuse computation from the shortest paths and trajectories computed in the previous time step which involve all detections up to time $t-1$. This is possible as in the online setting, the new network contains the previous network and some additional edges/nodes. It is important to note that a na\"ive extension of trajectories would violate optimality if the new evidence (\ie, detections at time $t$) changes the optimal trajectories for detections in previous time steps ($1, \cdots, t-1$). 

\begin{figure*}[t]
\subfloat[Original Network, $t=3$\label{fig:odDijkstra:a}]{
  \includegraphics[width=0.2\textwidth]{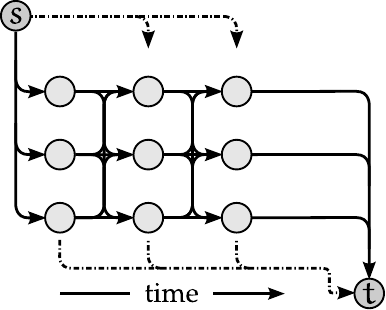}}\qquad
\subfloat[Solution, $t=3$\label{fig:odDijkstra:b}]{
  \includegraphics[width=0.2\textwidth]{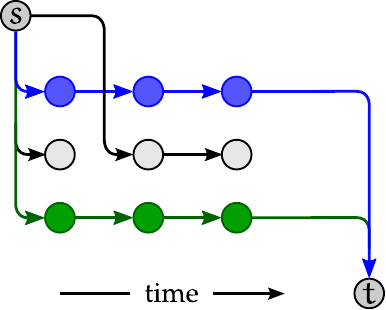}}\qquad
\subfloat[Nodes added for $t=4$\label{fig:odDijkstra:c}]{
  \includegraphics[width=0.2\textwidth]{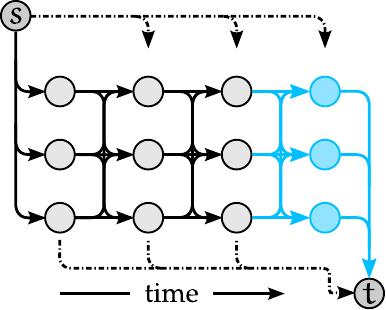}}\qquad
\subfloat[Solution in DAG\label{fig:odDijkstra:d}]{
  \includegraphics[width=0.2\textwidth]{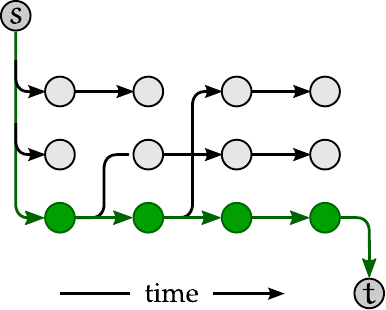}}\\
\subfloat[Merge Solution+Cache\label{fig:odDijkstra:e}]{
  \includegraphics[width=0.2\textwidth]{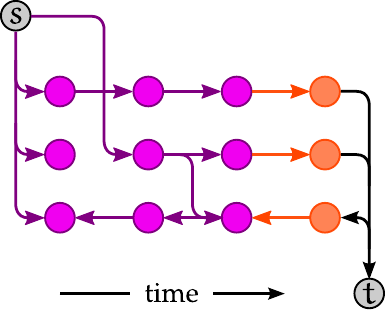}}\qquad
\subfloat[Broadcast Changes\label{fig:odDijkstra:f}]{
  \includegraphics[width=0.2\textwidth]{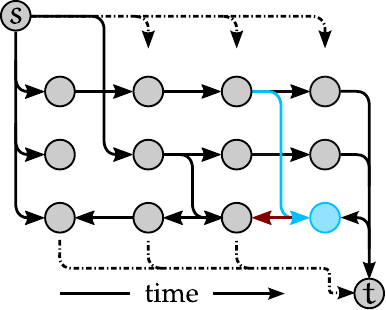}}\qquad
\subfloat[Decoded Trajectories\label{fig:odDijkstra:g}]{
  \includegraphics[width=0.2\textwidth]{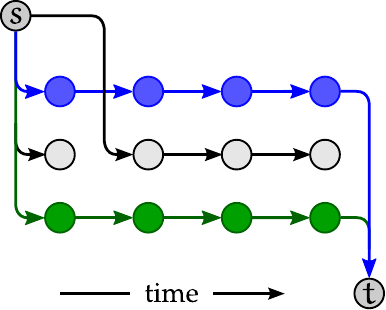}}\qquad
\subfloat[Clip Graph (mbodSSP)\label{fig:odDijkstra:h}]{
  \includegraphics[width=0.2\textwidth]{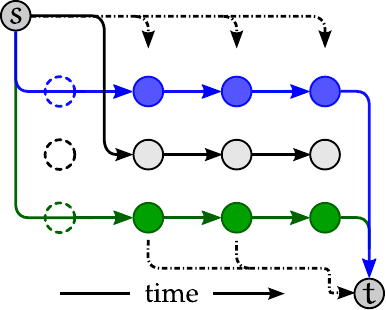}}
 \caption{{\bf Online \protect\subref{fig:odDijkstra:a}-\protect\subref{fig:odDijkstra:g} and Memory Bounded  \protect\subref
          {fig:odDijkstra:h} Algorithm:} 
          Assume that for the original network at $t=3$ \protect\subref{fig:odDijkstra:a}, the solution \protect\subref{fig:odDijkstra:b} is already known (shortest paths in blue and green). 
          A new set of detections (cyan nodes) for $t=4$ arrives and is connected to the graph \protect\subref{fig:odDijkstra:c} resulting in a DAG for which the shortest path can be found by applying one step of the DAG-SP algorithm \protect\subref{fig:odDijkstra:d}. 
          The predecessor maps from the previous (magenta nodes) and current timte step (orange nodes) are merged \protect\subref{fig:odDijkstra:e}. The queue initially contains nodes with outdated predecessors from the last shortest path (cyan).
          The priority queue is processed until convergence (\figref{fig:dDijkstra}) and the predecessors are updated \protect\subref{fig:odDijkstra:f}. 
          In this example, the algorithm terminates after one iteration and the optimal solution is found \protect\subref{fig:odDijkstra:g}. 
          For the mbodSSP algorithm, paths which pass through nodes outside the history window are merged with the entry costs of the next time step \protect\subref{fig:odDijkstra:h}.
         }
 \label{fig:odDijkstra}
\vspace{0cm}
\end{figure*}

We illustrate our dynamic online algorithm with an example in  \figref{fig:odDijkstra}. Consider the network specified in \subfigref{fig:odDijkstra:a} where the most recent time step is $t=3$ and  the optimal set of trajectories and predecessor maps have been computed.  \subfigref{fig:odDijkstra:b} illustrates the shortest paths as well as the two optimal trajectories (in blue and green respectively) computed at time $t=3$. 
As a new frame arrives, new nodes (depicted in cyan in \subfigref{fig:odDijkstra:c} are added, extending the graph to the next time step. 
The first trajectory can be computed by applying  dynamic programing only to the edges involving the new nodes,  
as in a DAG  the predecessors do not change from previous time steps. 
For successive trajectories (next iterations of the SSP algorithm), the predecessor maps from the previous time steps can be reused only if  the trajectories under consideration are currently in the same order as  when applying SSP to the previous time frame $t-1$ (\ie, before we received the new observation). 
This will not be the case when we have competing trajectories with similar costs, as given the new evidence (new frame) the ordering of these trajectories is likely to change. 
To handle this case, we utilize a caching strategy which keeps all predecessor maps for a cache   of $|\cc|$ frames in memory. Note that the dynamic broadcast algorithm described in Section \ref{subsec:dDijkstra} is  used within each iteration of SSP to obtain the predecessors with minimal computation. 

Coming back to our example, the optimal solution for the first iteration of SSP is found in \subfigref{fig:odDijkstra:d}. The predecessor maps for the new nodes (orange nodes in \subfigref{fig:odDijkstra:e}) have to be re-estimated only if they were part of the shortest path. Towards this goal, we employ our dynamic broadcasting strategy  where  only the node in cyan in \subfigref{fig:odDijkstra:f} is  put into the initial priority queue. 
As in this example the optimal trajectories are consistent with the ones from the previous time instant (see \subfigref{fig:odDijkstra:b} and \subfigref{fig:odDijkstra:g}), the optimal trajectories can be computed very efficiently, resulting in massive computational gains.

\subsection{Memory-Bounded Solution (mbodSSP)}\label{sec:mbodDijkstra}

While the odSSP algorithm handles the online setting, it does not scale to  very large problems since messages might broadcast back to very early frames in order to guarantee optimality. Furthermore, memory requirements still grow unbounded  with the length of the sequence. 
Thus,  these algorithms are not applicable to autonomous driving or surveillance scenarios.

In this section we propose a memory and computationally bounded approximation which we call ``mbodSSP''. 
The key intuition is that given some computation and memory budget, we can safely neglect  most of the past, and only retain the interesting information, \ie, the trajectories that were optimal. 
We refer the reader to \algref{algo:ssp_online_mbodDijkstra} for a depiction of the algorithm, and focus on explaining it through an example. 
Consider the case where we assume a budget of $\tau=4$ frames, and   the solution from the previous time step is given in \subfigref{fig:odDijkstra:g}. Before adding new nodes to the graph for the next time step $t+1$, we remove the oldest time step $t-\tau$ from the graph as illustrated in \subfigref{fig:odDijkstra:h} to maintain the memory/computation constraints.
Simply deleting edges from the graph is suboptimal as it completely discards computations from  previous time steps (before the new observation arrives).
In order to ``remember'' known paths, for each trajectory we sum the cost of the predecessor node at time $t-\tau$ to the  corresponding successor at time $t-\tau+1$. 
This strategy allows us to use the cache and remember previously found paths.
In the rare event that a trajectory which is not represented in the current cache is found,  we resort to odSSP on window $[t-\tau+1, t+1]$, guaranteeing a valid cache for the current SSP iteration. 
The cached predecessor maps are clipped using the same strategy. 
While optimality is violated as no changes can be applied to paths beyond $t-\tau+1$, our experiments demonstrate that for many cases of practical interest small windows are sufficient to obtain near-optimal solutions. In particular, our memory bounded approximation is able to maintain track ids over periods much longer than the window itself. In contrast, when splitting the sequence into batches and applying the min-cost flow network algorithm separately to each batch, ad-hoc heuristics are required to resolve this problem.

%%%%%%%%%%%%%%%%%%%%%%%%%%%%%%%%%%%%%%%%%%%%%%%%%%%%%%%%%%%%%%%%%%%%%%%%%%%%%%%%
% SSP-mbodDijkstra ALGORITHM
\begin{algorithm}[t]
{\small
  \DontPrintSemicolon
  \KwIn{Current Detections $\mathcal{X}{_{\tmax}} = \{\bx_{\tmax}\}$,\linebreak Graph $G(V,E,C,f,s,t)$, Cache $\cal C$}
  \KwOut{Set of trajectory hypotheses $\mathcal{T} = \{T_k\}\;$}
%   \tcp{clip graph at the beginning (mbodDijkstra)}
  \If{memorybounded}{
    \ForEach{node u $\in [\tmax-\tau]$}{
    \tcp{\hspace*{-0.3em}remember predecessor by updating $c_{en,i}$}
    $G(f) \gets$ UpdateEntryEdge($G(f),u$)\;
    $G(f) \gets $ RemoveObservation($G(f),u)$ %\tcp*{remove edges}
    }
  }
%   \tcp{From here on: mbodDijkstra $\equiv$ odDijkstra}
  %\tcp{add arriving observations to graph}
  $G(f) \shortgets$ AddObservations($G(f),\mathcal{X}{_{\tmax}}$) \tcp*{\hspace*{-0.4em} still a DAG}
  %$f(G) \gets 0 \;\forall\, \cx_{t{_i}}$ \tcp*{initialize flow for new detections}
%  \tcp{take $\pi^{(0)}$ for the DAG from last time step}
  $\pi^{(0)} \gets {\cal C}^{(0)}(\tmax-1)$\; 
  \tcp{run DAG-SP for edges $(u,v) \in [\tmax-1,\tmax]$}
  $\gamma^{(0)}, \pi^{(0)} \gets$ DAG-SP($G_r(f),\tmax-1$) \;
  $G_r^{(0)}(f) \hspace*{-0.1em}\gets\hspace*{-0.1em}$ ConvertEdgeCosts($G(f), \gamma^{(0)}, \pi^{(0)}, \tmax-1$)\; %\tcp*{initialize $G_r(f)$} 
  $G^{(1)}_r(f) \hspace*{-0.15em}\gets\hspace*{-0.15em}$ ComputeResidualGraph($G^{(0)}_r(f),\pi^{(0)}$)\;
  $q \gets \emptyset$,\, %\tcp*{$q$ is maintained for every iteration $k$}
  %
  %\tcp{find $k$-th shortest paths}
  $k \gets 0$\;
  \While{$1$} {
    $k \gets k+1$\;
    \tcp{$\gamma^{(k-1)}_{\tmax-\delta_i} = \gamma^{(k-1)}_{\tmax}$, $\delta_i \in \{0,\ldots,|\cc|\}$ }
    ${\delta_i} \gets$ MostRecentCache($\mathcal{C}, \gamma^{(l)}\; \forall\; l=0,\ldots,k-1$) \;
    $\pi^{(k)} \gets \mathcal{C}(\delta_i,k)$ \tcp*{update predecessor map $\pi_k$}
    %\tcp{predecessors for $\gamma^{(k-1)}_{\tmax-\delta_i} \setminus \gamma^{(k-1)}_{\tmax}$ are invalid}
%    \tcp{invalid predecessors}
    $q \gets \{\gamma^{(k-1)}_\tmax, (u,v) \in [\tmax-\delta_i,\tmax]\}$\; 
    \tcp{\algref{algo:ssp_offline_dDijkstra}, line 10}
    $G_r^{(k+1)}(f), \gamma^{(k)} \gets$ ProcessQueue($q$, $\pi^{(k)}$, $G_r^{(k)}(f)$)\; 
    }
    \Return{$\mathcal{T}$}\;
  \caption{mbodSSP}
  \label{algo:ssp_online_mbodDijkstra}
}
\end{algorithm}

%%%%%%%%%%%%%%%%%%%%%%%%%%%%%%%%%%%%%%%%%%%%%%%%%%%%%%%%%%%%%%%%%%%%%%%%%%%%%%%%

%% file: results.tex
\section{Experimental Evaluation}
\label{sec:results}

We evaluate our algorithms on the challenging KITTI \cite{Geiger2012CVPR} and PETS 2009 \cite{Ferryman2009PETS} tracking benchmarks. For PETS we use the object detections provided by Andriyenko \etal \cite{Andriyenko2012CVPR}. For KITTI we compare the DPM reference detections \cite{Felzenszwalb2010PAMI} provided on the KITTI website\footnote{\url{http://www.cvlibs.net/datasets/kitti/eval\_tracking.php}} with the recently proposed Regionlet detections \cite{Wang2013ICCV} provided to us by the authors.

We convert the detection score of each bounding box $d_i$ into a unary cost value $C_i$ using logistic regression $C_i = 1 / (1 + \mathrm e^{\beta d_i})$ on the training set.
To encode association costs, we use six different pairwise similarity features $\mathbf{s} = \{s_l\}$: bounding box overlap, orientation similarity, color histogram similarity, cross-correlation, location similarity, and optical flow overlap.
Similar to the detection scores we pass the association features through a logistic function using logistic regression yielding $\mathbf{s}\in[0,1]^6$.
The detection/association cost for each edge $(u,v)$ is then defined as $C_{u,v} = ((\mathbf{1}-\bs)+\bo)^\top \bw$, 
where $\bo$ denotes an offset and $\bw$ the scale. Note that the offset is required to allow for negative as well as positive costs.
All parameters ($\bo, \bw$) have been obtained using block coordinate descent on the respective training sets and kept fix during all our experiments. 
We refer the reader to the supplementary material for further details on the parameter setting, additional results, and videos. 

\paragraph{Comparison to State-of-the-art on KITTI:}
We first compare the proposed dSSP and mbodSSP algorithms against four state-of-the-art baselines \cite{Geiger2014PAMI,Pirsiavash2011CVPR,Andriyenko2012CVPR,Milan2014PAMI} as well as the pairwise optimal Hungarian method (``HM'') on the challenging KITTI dataset using the DPM reference detections \cite{Felzenszwalb2010PAMI}. The metrics we use are described in \cite{Li2009CVPRa,Bernardin2008JIVP}. 
As shown in Table~\ref{table:compare_tracking_KITTI} (left part), the optimal algorithm (``SSP'') outperforms all other methods. Note that all discussed optimal batch solvers obtain identical tracking results, but with different run time thus we only state results for SSP. In our experiments, we made use of a relatively low threshold $d_i=-0.3$ for the object detector to avoid early pruning and evaluate each method with respect to outlier rejection performance. Note that our method attains the best performance with respect to mostly tracked trajectories (``MT'') while only exhibiting a slightly higher false alarm rate (``FAR'') than the other methods.
The method of \cite{Andriyenko2012CVPR} struggled with the presence of outliers and $d_i=0.0$ was used to obtain meaningful results. 
Also note the little loss in performance when running mbodSSP for a window length of $\tau=10$.
Compared to the non-optimal DP solution \cite{Pirsiavash2011CVPR}, mbodSSP achieves higher performance, especially in terms of identity switches and fragmentations. We provide additional qualitative results in the supplementary material.
We also experiment with the detector of \cite{Wang2013ICCV}, which yields better results  particularly for occluded objects. The increasing tracking performance indicates that with a good enough detector our approach could be used in practical applications. 

\begin{figure*}[t!]
  \centering
 \subfloat[Run Time\label{fig:runtime_memory:runtime}]{
     \includegraphics[width=0.32\textwidth]{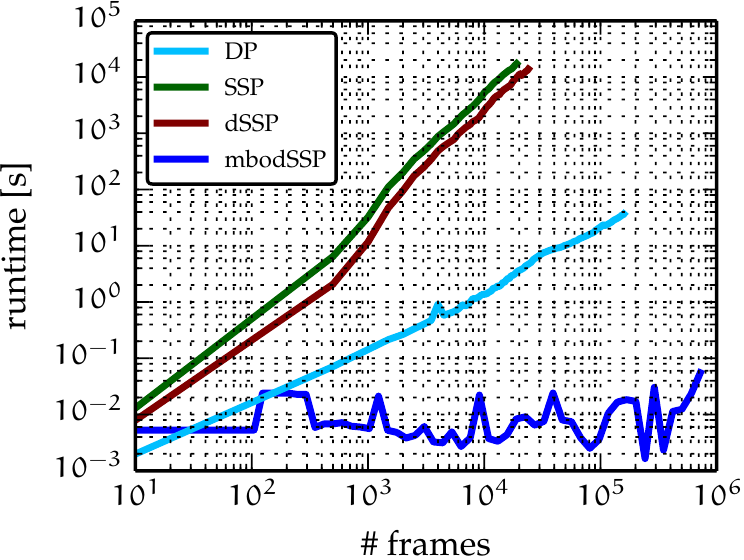}}\hfill
 \subfloat[Memory Consumption\label{fig:runtime_memory:memory}]{
     \includegraphics[width=0.32\textwidth]{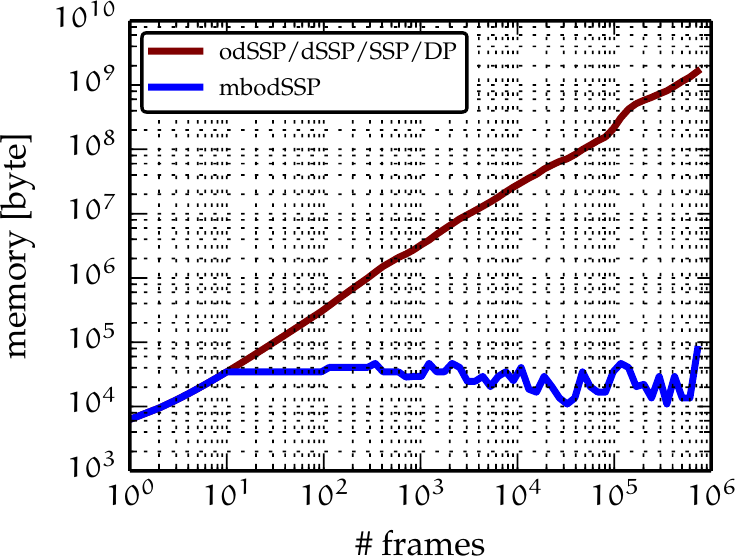}}\hfill
 \subfloat[Impact of History Length $\tau$\label{fig:runtime_memory:history}]{
     \includegraphics[width=0.32\textwidth]{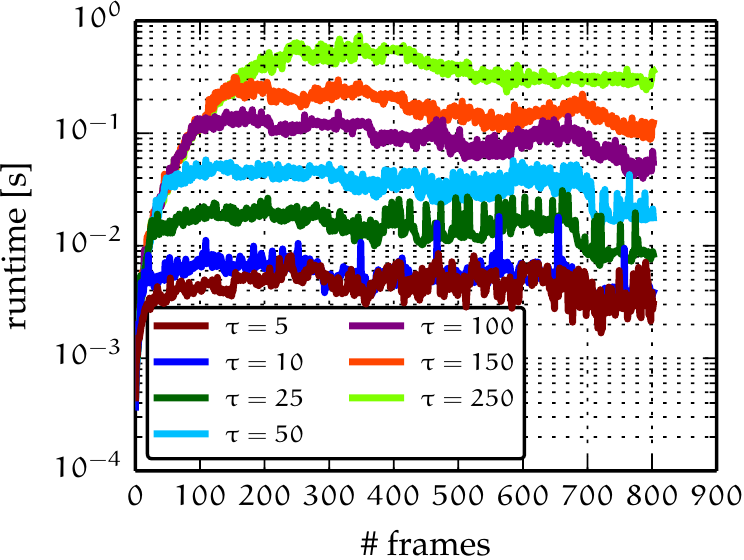}}\\[-0.7em]
  \caption{\textbf{Run Time and Memory Comparison.} 
           We compare computational performance of all solvers using one long sequence without ground truth annotations. This figure shows the mean runtime \protect\subref{fig:runtime_memory:runtime} and idealized memory consumption \protect\subref{fig:runtime_memory:memory} for every solver. Additionally, we show the impact of different values for the history length $\tau$ \protect\subref{fig:runtime_memory:history}.
          }
  \label{fig:runtime_memory}
\vspace{0cm}
\end{figure*}

\begin{table}\setlength{\tabcolsep}{3pt}
\scalebox{0.75}{\input{metrics_raw_kitti_testing_car.tex}}
\caption{Comparison of our proposed methods to four state of the art methods and a HM baseline implementation on KITTI-Car  using the DPM reference detections and Regionlet detections (marked with a star).}
\label{table:compare_tracking_KITTI}
\vspace{0.2cm}
\end{table}

\begin{table}\setlength{\tabcolsep}{7.5pt}
\scalebox{0.75}{\input{metrics_raw_PETS2009_pedestrian.tex}}
  \caption{Comparison of our proposed method to three baselines on PETS 2009 sequence ``S2.L1''.
}
  \label{table:compare_tracking_PETS}
\vspace{0.25cm}
\end{table}

\paragraph{Comparison to  State-of-the-art on PETS2009:}
We additionally evaluate our methods on the commonly used PETS2009 dataset for sequence ``S2.L1''. We used the detections and ground truth provided by the authors of \cite{Andriyenko2011CVPR,Andriyenko2012CVPR,Milan2014PAMI}. Both, the optimal algorithms and the memory-bounded approximation perform on par with current state-of-the-art. In particular performance for precision-based measures is notably good. Note that we used the same parameters as for the results presented on KITTI.

\paragraph{Comparing Min Cost Flow Solutions:}
We compare our globally optimal methods (dSSP, odSSP) as well as our approximate mbodSSP algorithm for window size $\tau=10$ against a regular SSP implementation using Dijkstra's algorithm (as described in \cite{Cormen2001}) as well as the non-optimal DP solution of\cite{Pirsiavash2011CVPR} in terms of run time and memory consumption. 
For a fair comparison, we implement all our solvers in Python using the same data structures. 
\figref{fig:runtime_memory} \subref{fig:runtime_memory:runtime}-\subref{fig:runtime_memory:memory} depict execution time and memory consumption as a function of the number of frames for a very long sequence on KITTI. 
Note that our globally optimal dynamic solver (dSSP) outperforms the regular Dijkstra implementation (SSP) by a factor of 3 on average.  
Importantly, our experiments validate that the time complexity of mbodSSP is independent of the sequence length. Thus, it outperforms DP \cite{Pirsiavash2011CVPR} in memory, computation as well as accuracy (see Table \ref{table:compare_tracking_KITTI}).
A run time evaluation for  KITTI  including a comparison for the different training scenarios is discussed in greater depth in the supplementary material.

\paragraph{Sliding Window Size of mbodSSP:}
We evaluate the tracking performance of mbodSSP for different values of $\tau$  on  KITTI. 
As shown in \subfigref{fig:runtime_memory:history},  for a value of $\tau=10$, our non-optimized Python implementation of mbodDijkstra requires less than 10ms which is sufficient for many real-time online applications.

\paragraph{Supplementary Material:}
In the supplementary material we give an overview of related (SSP) algorithms, specify the details of our association features, provide an analysis of the robustness of the proposed model against variation of the parameters, and show additional results in terms of runtime and performance. We also encourage the reviewers to watch the supplementary video which illustrates and compares our tracking results for several KITTI sequences.

%% file: metrics_raw_kitti_testing_car.tex
\begin{tabular}{lccccc|cc||cc}
\hline
 &HM &\cite{Andriyenko2012CVPR} &\cite{Milan2014PAMI} &\cite{Pirsiavash2011CVPR} &\cite{Geiger2014PAMI} &mbodSSP &SSP &mbodSSP* &SSP* \\%\midrule
\hline
{MOTA} & {0.42} & {0.35} & {0.48} & {0.44} & {0.52} & {0.52} & \textbf{0.54} & {0.67} & {0.67}\\
{MOTP} & {0.78} & {0.75} & {0.77} & {0.78} & {0.78} & {0.78} & {0.78} & {0.79} & {0.79}\\
% {MODA} & {0.42} & {0.36} & {0.48} & {0.52} & {0.52} & {0.52} & \textbf{0.54} & {0.67} & {0.67}\\
% {MODP} & {0.53} & {0.54} & {0.58} & {0.54} & {0.58} & \textbf{0.59} & \textbf{0.59} & {0.66} & {\textbf{0.67}}\\
% {Recall} & {0.43} & {0.50} & {0.54} & {0.46} & {0.54} & \textbf{0.56} & \textbf{0.58} & {0.78} & {\textbf{0.80}}\\
% {Prec.} & {\textbf{0.97}} & {0.77} & {0.90} & {0.96} & {0.95} & {0.93} & {0.94} & {0.88} & {0.87}\\
{F1} & {0.60} & {0.61} & {0.67} & {0.62} & {0.69} & \textbf{0.70} & \textbf{0.71} & {0.83} & {0.83}\\
{FAR} & {\textbf{0.048}} & {0.46} & {0.18} & {0.053} & {0.083} & {0.14} & {0.11} & {0.34} & {0.40}\\
{MT} & {0.077} & {0.11} & {0.14} & {0.11} & {0.14} & \textbf{0.15} & \textbf{0.21} & {0.34} & {0.41}\\
% {PT} & {0.50} & {\textbf{0.56}} & {0.52} & {0.49} & {0.52} & {0.55} & {0.51} & {0.55} & {0.51}\\
{ML} & {0.42} & {0.34} & {0.34} & {0.39} & {0.35} & \textbf{0.30} & \textbf{0.27} & {0.10} & {0.090}\\
% {TP} & {15293} & {17734} & {19057} & {16113} & {19308} & \textbf{19909} & \textbf{20447} & {28623} & {\textbf{29714}}\\
% {FP} & {\textbf{535}} & {5161} & {2045} & {588} & {923} & {1588} & {1279} & {3765} & {4500}\\
% {Misses} & {20121} & {17733} & {16401} & {19287} & {16139} & \textbf{15506} & \textbf{15034} & {8022} & {\textbf{7279}}\\
{IDS} & {12} & {223} & {125} & {2738} & {33} & {\textbf{0}} & \textbf{7} & {117} & {194}\\
{Frag.} & {578} & {624} & {\textbf{401}} & {3241} & {540} & {708} & {717} & {894} & {977}\\
% {Tracker Objects} & {16532} & {25860} & {23188} & {17493} & {21945} & {23795} & {23957} & {35950} & {38393}\\
% {Tracker Trajectories} & {1102} & {824} & {935} & {458} & {1049} & {1653} & {1135} & {2520} & {2485}\\
%\bottomrule
\hline
\end{tabular}

%% file: metrics_raw_PETS2009_pedestrian.tex
\begin{tabular}{lcccc|cc}
% \toprule
\hline
 & \cite{Andriyenko2011CVPR} &\cite{Andriyenko2012CVPR} & EKF \cite{Milan2014PAMI} & \cite{Milan2014PAMI} &mbodSSP &SSP \\\hline
{MOTA}          & {0.81} & \textbf{0.96} & {0.68} & {0.91} & {0.89}   & {0.91}\\
{MOTP}          & {0.76} & {0.79} & {0.77} & {0.80} &\textbf{0.87}   & \textbf{0.87}\\
% {MODA}          & {n/a}  & {n/a}  & {n/a}  & {n/a}  &{0.89}   & {0.91}\\
% {MODP}          & {n/a}  & {n/a}  & {n/a}  & {n/a}  &{0.87}   & {0.87}\\
% {Recall}        & {n/a}  & {n/a}  & {0.70} & \textbf{0.92} &{0.90}   & \textbf{{0.92}}\\
% {Precision}     & {n/a}  & {n/a}  & {0.98} & {0.98} &\textbf{0.99}   & \textbf{0.99}\\
% {F1}            & {n/a}  & {n/a}  & {0.82} & \textbf{0.95} &{0.94}   & \textbf{{0.95}}\\
% {FAR}           & {n/a}  & {n/a}  & {0.08} & {0.07} &\textbf{0.057}  & {0.067}\\
{MT}            & {0.83} & \textbf{0.96} & {0.39} & {0.91} &{0.89}   & {0.89}\\
{PT}            & \textbf{0.17} & {0.04} & {0.57} & {0.05} &{0.11}   & {0.11}\\
{ML}            & \textbf{0.0}  & \textbf{0.0}  & {0.04} & {0.04} &\textbf{0.0}    & \textbf{0.0}\\
% {TP}            & {n/a}  & {n/a}  & {n/a}  & {n/a}  &{4153}   & {4247}\\
% {FP}            & {n/a}  & {n/a}  & {65}   & {59}   &\textbf{45}     & {53}\\
% {Misses}        & {n/a}  & {n/a}  & {1173} & \textbf{302}  &{461}    & {367}\\
{ID-switches}   & {15}   & {10}   & {25}   & {11}   &\textbf{7}      & {23}\\
{Fragmentations}& {21}   & {8}    & {30}   & \textbf{6}    &{100}    & {100}\\
% {Tracker Objects} & {n/a} & {n/a} & {n/a}  & {n/a}  &{4198}   & {4300}\\
% {Tracker Trajectories} & {n/a} & {n/a} & {n/a} & {n/a} & {103} & {101}\\
\hline
% \bottomrule
\end{tabular}

%% file: conc.tex
\section{Conclusions}

In this paper we have proposed solutions to make the use of min-cost flow  tracking by detection possible in real world scenarios. Towards this goal we have designed algorithms that are dynamic, can handle an online data stream and are bounded in memory and computation. 
We have demonstrated the performance of our algorithms in challenging autonomous driving and surveillance scenarios. 
In future work we plan to extend our approach to explicitly tackle long-term occlusion as well as to  incorporate additional high-level features, \eg, map information.  